\documentclass[letterpaper, 10 pt, journal, twoside]{IEEEtran}
\ifCLASSINFOpdf
\else
\fi
\usepackage{amsmath,amsfonts}
\usepackage{algorithm}
\usepackage{algpseudocode}
\usepackage{array}
\usepackage[font=scriptsize,caption=false]{subfig}
\usepackage{textcomp}
\usepackage{stfloats}
\usepackage{url}
\usepackage{verbatim}
\usepackage{graphicx}
\usepackage{soul}

\usepackage{hyperref}

\usepackage{cite}
\usepackage{xcolor,pifont}
\newcommand*\colourcheck[1]{%
  \expandafter\newcommand\csname #1check\endcsname{\textcolor{#1}{\ding{52}}}%
}
\newcommand*\colorcross[1]{%
  \expandafter\newcommand\csname #1cross\endcsname{\textcolor{#1}{\ding{56}}}%
}
\let\labelindent\relax
\usepackage[inline]{enumitem}

\colourcheck{green}
\colorcross{red}

\newcommand{\TaskTree}{\mathcal{T}}

\newcommand{\Ingredients}{\mathbb{I}}
\newcommand{\DishType}{\mathcal{D}}
\newcommand{\Candidates}{\mathsf{candidates}}
\newcommand{\Subgraph}{\mathsf{subgraph}}
\newcommand{\Goal}{\mathcal{G}}

\newcommand{\SetS}{\mathbb{S}}

\newcommand{\Score}{\mathsf{score}}

\newcommand{\C}{\mathsf{C}}

\begin{document}
%
\title{Approximate Task Tree Retrieval in a Knowledge Network for Robotic Cooking}
%
%
%

\author{Md. Sadman Sakib$^{1}$, David Paulius$^{2}$, and Yu Sun$^{1}$%
\thanks{Manuscript received: February 24, 2022; Revised May 20, 2022; Accepted June 25, 2022.}
\thanks{This paper was recommended for publication by Jingang Yi upon evaluation of the Associate Editor and Reviewers' comments.
This work was supported by the National Science Foundation under Grants 1910040 and 1812933.
} 
\thanks{$^{1}$Md Sadman Sakib and Yu Sun are members of the Robot Perception and Action Lab (RPAL), which is part of the Department of Computer Science \& Engineering at the University of South Florida, Tampa, FL, USA.
        {\tt\footnotesize mdsadman@usf.edu, yusun@usf.edu}}%
\thanks{$^{2} $David Paulius is a postdoctoral researcher in the Intelligent Robot Lab at Brown University, Providence, RI, USA.
        {\tt\footnotesize dpaulius@cs.brown.edu}}%
\thanks{Digital Object Identifier (DOI): see top of this page.}
}
%
%

\markboth{IEEE Robotics and Automation Letters. Preprint Version. Accepted June~2022}
{Sakib \MakeLowercase{\textit{et al.}}: Approximate Task Tree Retrieval} 

%



\maketitle

\begin{abstract}
Flexible task planning continues to pose a difficult challenge for robots, where a robot is unable to creatively adapt their task plans to new or unseen problems, which is mainly due to the limited knowledge it has about its actions and world.
Motivated by a human's ability to adapt, we explore how task plans from a knowledge graph, known as the 
\emph{Functional Object-Oriented Network}
(FOON), can be generated for novel problems requiring concepts that are not readily available to the robot in its knowledge base.
Knowledge from 140 cooking recipes are structured in a FOON knowledge graph, which is used for acquiring task plan sequences known as task trees.
Task trees can be modified to replicate recipes in a FOON knowledge graph format, which can be useful for enriching FOON with new recipes containing unknown object and state combinations, by relying upon semantic similarity.
We demonstrate the power of task tree generation to create task trees with never-before-seen ingredient and state combinations as seen in recipes from the Recipe1M+ dataset, with which we evaluate the quality of the trees based on how accurately they depict newly added ingredients.
Our experimental results show that our system is able to provide task sequences with 76\%
correctness. 
\end{abstract}


%
\IEEEpeerreviewmaketitle

%
%
%
%

%
\section{Introduction}
\IEEEPARstart{M}{ajor} efforts in robotics research have been devoted to the development of intelligent agents with the ability to understand human intentions and to perform actions for problems in human-centered domains.
Such domains include, but are not limited to, the application of robots for assisting the elderly and disabled, food delivery, and cooking activities.
However, the main difficulty in designing robots for human-centered domains lies in the variety of tasks and the dynamic nature of environments in which these robots will operate.

In terms of robotic cooking, ingredients or objects may come in different forms, shapes or sizes, and there are many states to account for when executing recipes~\cite{jelodar2019joint}.
Additionally, there may be times where a robot is unable to execute an entire recipe due to the unavailability of certain ingredients or objects in its surroundings; this may happen when a robot is required to prepare meal variations. 
In the latter case, a robot would need to adapt its knowledge to include or disregard ingredients, which may or may not be known by the robot, in its task plan, which is not often considered in modern robot applications.
Motivated by a human's ability to creatively adapt to novel scenarios, we explore the problem of generating task plans for novel scenarios that would be presented to a robot, given the limited knowledge available to it as a knowledge graph.

\begin{figure}[t]
	\centering
	\includegraphics[width=0.8\columnwidth]{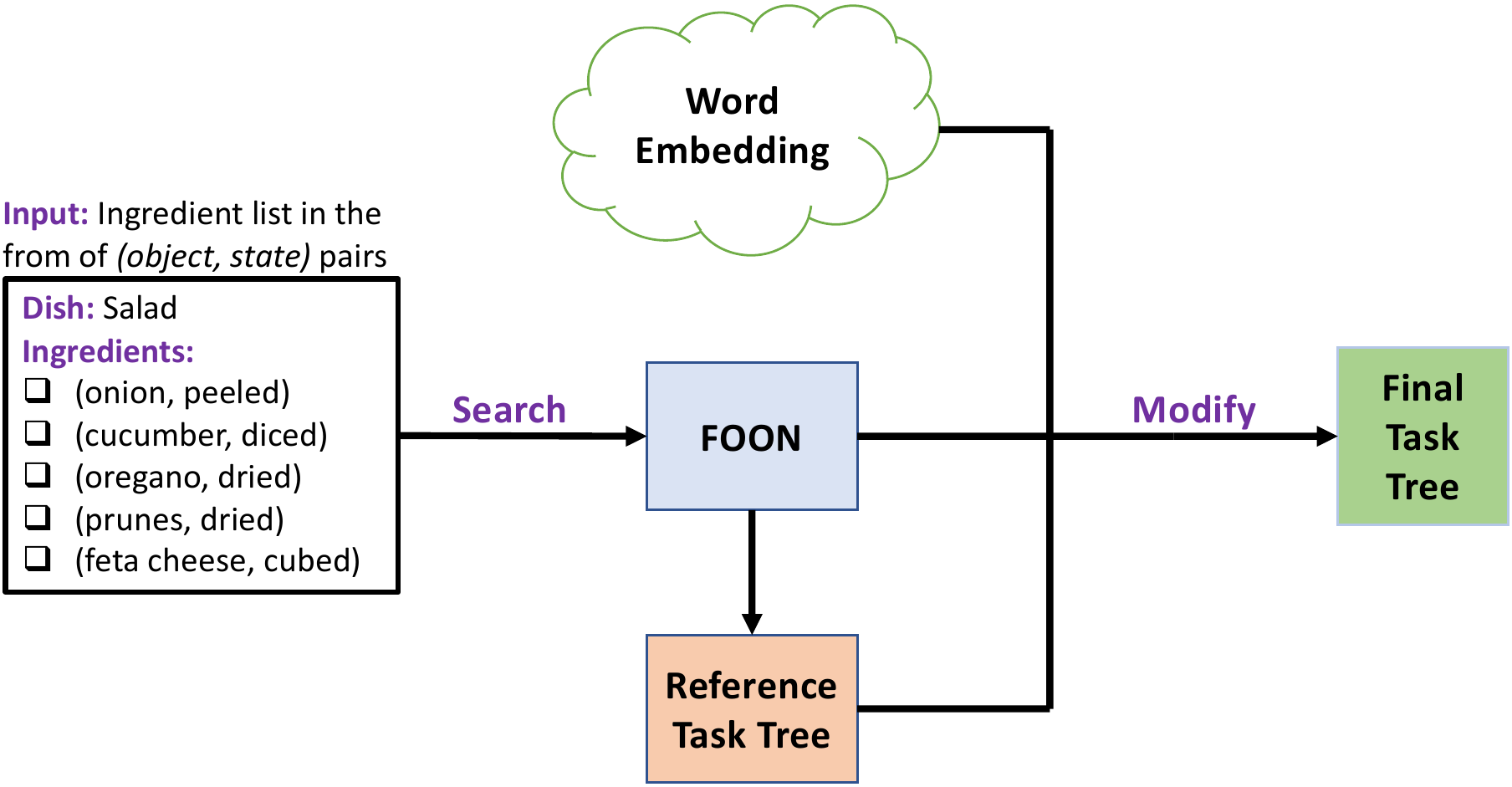}
	\caption{Overview of our task tree generation procedure for novel problems using FOON. As input to our task tree generation system, a set of ingredients and their states are given, from which a novel task tree can be generated.}
	\label{fig:pipeline3}
	\vspace{-0.15in}
\end{figure}

In this work, we use a knowledge representation known as the 
\emph{Functional Object-Oriented Network} (FOON)~\cite{paulius2016functional}, which builds upon previous work on joint object-action representation~\cite{Ren2013,SunRAS2013,Lin2015a}.
%
%
%
In prior work, we introduced how a FOON can be created from video annotations and used for task planning~\cite{paulius2016functional}. 
%
However, as with other works, task planning is limited to the knowledge found in a FOON, as it only contains knowledge for a limited number of recipe and ingredient variations.
For example, if a robot were to prepare a salad with a certain combination of ingredients that have never been encountered together before in FOON, then this would pose a problem since there is no concept of that type of salad.
Previously, we investigated how the knowledge in FOON can be extended so as to generalize concepts across object types~\cite{paulius2018functional}.
Based on this intuition, we propose that existing knowledge of similar recipes can be used to derive alternative solutions (as graphs) that did not exist before.
Therefore, in this work, we introduce a procedure known as \textit{task tree generation} (described in Figure~\ref{fig:pipeline3}), where a \textit{reference task tree} is extracted from FOON and then modified to match a required set of ingredients for a given recipe.
This may require the addition of possibly never-before-seen ingredients and the removal of ingredients in the reference task tree that are not valid for the new recipe.
To show the effectiveness of our approach, we apply task tree generation on recipes from the Recipe1M+ dataset~\cite{marin2019learning}.
This is particularly useful for creating new FOONs and task plans for never-before-seen recipes without manual effort.

Our contributions and outcomes in this paper are as follows:
\begin{enumerate*}[label=(\roman*)]
\item We develop a recipe generation pipeline that takes a set of ingredients as input and outputs a task plan in the form of a task tree;
\item We design a heuristic-based search algorithm to find a recipe from the knowledge graph, which closely matches with the required ingredients, without exploring all possible paths and thus minimizing time and space cost;
\item  We present a task tree visualization method that highlights how each ingredient's state changes in a recipe known as the \textit{progress line}; and 
\item We perform an evaluation on the generalizability of FOON for Recipe1M+ recipes.
\end{enumerate*}

\section{Background}

\subsection{Functional Object-Oriented Network}
\label{sec:FOON}
\begin{figure}[!t]
	\centering
    \includegraphics[width=0.9\columnwidth]{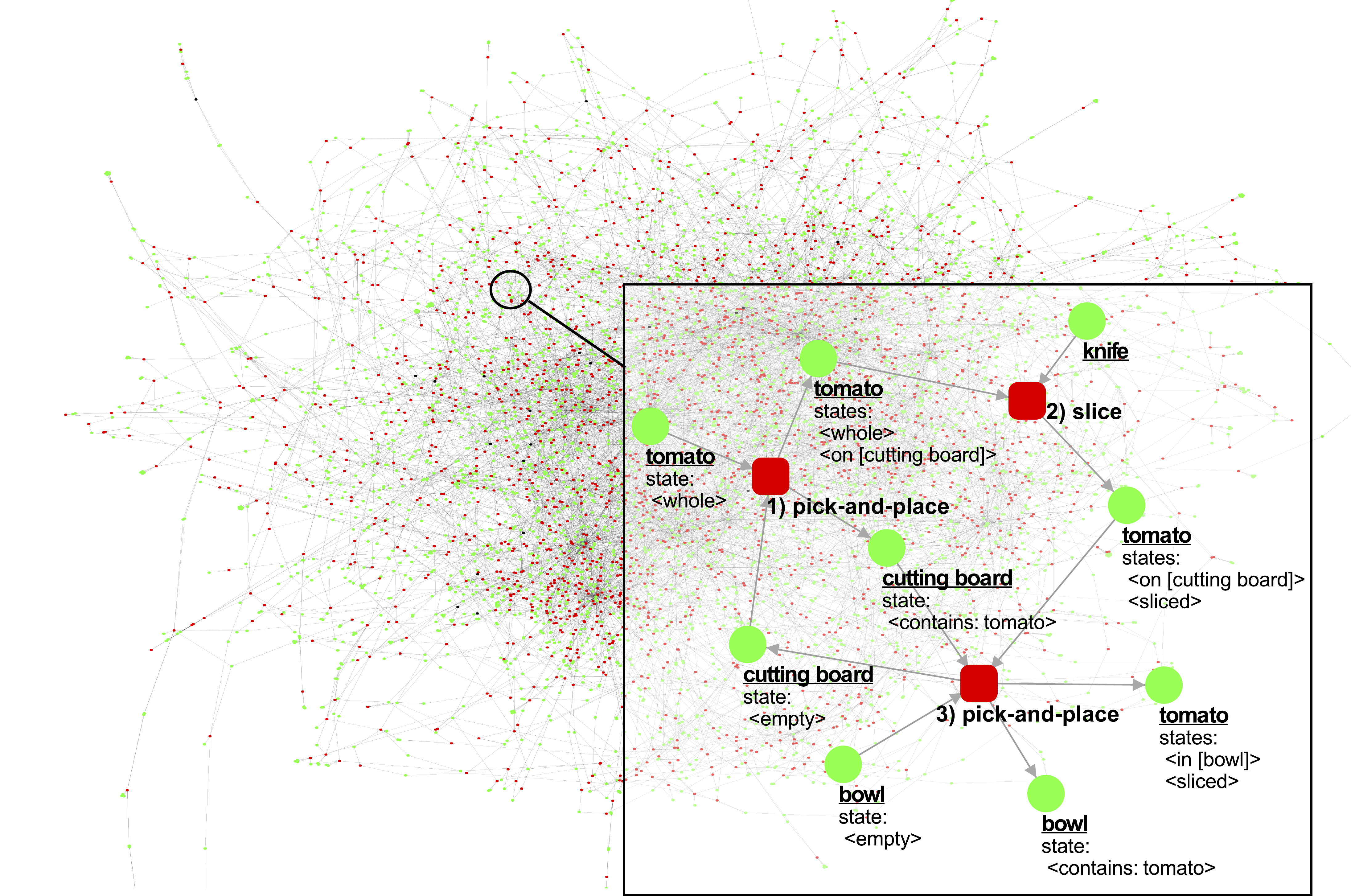}
	\caption{
	Illustration of a universal FOON made of 140 recipe subgraphs, and three (3) connected functional units, where object nodes are denoted by green circles and motion nodes are denoted by red squares.
	}
	\label{fig:unit}
	\vspace{-0.2in}
\end{figure}

A FOON is a bipartite network with two types of nodes: \textit{object nodes} and \textit{motion nodes}.
Affordances~\cite{Gibson_1977} are depicted via edges that connect objects to actions and enforce action sequencing.
A fundamental structure known as a \textit{functional unit} represents actions in FOON by describing the state change of objects before and after execution, and it has \textit{input} object nodes, \textit{output} object nodes, and a motion node.
It is akin to a planning operator in PDDL (short for \textit{Planning Domain Definition Language})
~\cite{mcdermott1998pddl}, where input and output nodes describe preconditions and effects respectively.
Figure~\ref{fig:unit} depicts examples of functional units for the action sequence of: 1) picking and placing a \textit{whole tomato} on a \textit{cutting board}, 2) slicing the \textit{tomato} with a \textit{knife}, and 3) pouring the \textit{sliced tomato} from the \textit{cutting board} into a \textit{bowl}.

Typically, we create FOONs by annotating video demonstrations.
A FOON representing a single activity is referred to as a \textit{subgraph}; a subgraph contains functional units in sequence to describe objects' states before and after each action occurs, and what objects are being manipulated.
Presently, this annotation process is done manually, but previous work has investigated how graphs can be annotated in a semi-automatic manner~\cite{jelodar2018long}.
Two or more subgraphs can be merged together to form what we call a {\it universal FOON}. 
A universal FOON can contain variations of recipes once it has been merged with several sources of knowledge. 
This merging procedure is simply a union operation applied to all functional units from each subgraph we wish to combine; as a result, duplicate functional units are eliminated in the merged network~\cite{paulius2016functional}.
Presently, FOON comprises of 140 subgraph annotations of videos from YouTube, Activity-Net~\cite{caba2015activitynet}, and EPIC-KITCHENS~\cite{Damen2018EPICKITCHENS}, which are all available on our website~\cite{foonet}.

\subsubsection{Task Planning with FOON}

Task planning involves retrieving a \textit{task tree}~\cite{paulius2016functional} (a high-level task plan) that satisfies a goal defined as an object node.
Finding a task tree requires knowledge about the state of a robot's environment, specifically what objects are available and the state(s) in which they are in.
Starting from the goal, the algorithm searches for candidate functional units in a depth-wise manner, and then, for each candidate unit, it searches its input nodes in a breadth-wise manner to see if they are available to the robot.
More recently, we introduced a modified version of the algorithm to find the optimal path to reach a goal node, which is reflected by weights given to each functional unit corresponding to its success rate of execution by a robot~\cite{paulius2021weighted}.
However, these algorithms are not suited to generate task trees for recipes not found in FOON; for this task, we use semantic similarity~\cite{paulius2018functional} while searching to modify task trees as required
for a given problem and ingredient set.

\subsection{Related Works}
Developing a robotic agent that understands human instructions like ``cut the onion in half" and executing them precisely in a real world setting is a very difficult task. The most noteworthy work to address this challenge is the knowledge processing framework {\sc{KnowRob}} \cite{beetz2009}. 
They created this knowledge base by collecting information through observations in a sensor-equipped environment. 
%
%
Although the robots in \cite{pancake} were successful in retrieving a pancake recipe and executing it using the {\sc{KnowRob}} reasoning framework, the authors did not show whether the robots are able to modify the recipe to meet users' preference. On the contrary, our work focuses on this degree of flexibility so that one can obtain a task plan for any combination of ingredients.  

Furniture assembly and cooking are similar in a sense that both of these task domains begin with raw materials and end with a finished product in the form of an assembled furniture piece or a dish. 
Given geometric details of parts, IkeaBot~\cite{ikeabot} deduces a task plan to assemble furniture by using geometric reasoning and checking possible sub-assemblies, which is computationally expensive. On the other hand, we explore the FOON knowledge base by using an efficient heuristic-based graph search. 
\cite{Lee2021IKEAFA} relied on deep learning whereas \cite{onto_ikea} created an ontology based on the product description found on IKEA website to retrieve information about an object. 
%
Manipulation action tree banks~\cite{yang2014manipulation} were proposed to organize manipulation actions in multiple levels of abstraction, which a robot can use for understanding and executing a task.
Unlike our work, they did not handle different object states and are unable to produce a plan for complex manipulation tasks. 
More recently, the authors in~\cite{Sera2021AssemblyPB} presented a method for robot assembly planning by recognizing graphical instruction manuals. They identify objects from a series of images, find the relationship between the previous and current image, and create a task sequence graph for robot execution.

Efforts have been made to dynamically modify a recipe to meet certain constraints.  \cite{Varshney2013ABD} gathered various datasets to understand existing recipes, which flavors people like, and chemical structure of ingredients to invent novel recipes based on big data approaches.   
\cite{Mller2017CookingME} proposed a new approach to generate novel recipes based on ingredients, preparation time, and cooking steps, where a user can specify those parameters. They used a taxonomy to define the semantics of ingredients and cooking steps where the inner nodes refer to generalized items (e.g., meat) and the items close to the leaf nodes are more specific (e.g., beef, pork). 
EvoChef~\cite{Jabeen2019EvoChefSM} and AutoChef~\cite{Jabeen2020AutoChefAG}, created novel recipes using natural language processing and evolutionary algorithms but without considering user preference.
\cite{lida1} optimize a recipe according to consumers' taste by tuning parameters such as cooking time, amount of salt. However, they maintain the same ingredients in their plans, whereas our work shows more flexibility by making the same dish using different ingredient combinations based on the consumers' dietary preferences. 


Some works focused on finding a reasonable substitution for an ingredient. \cite{DBLP:journals/corr/AchananuparpW16} collected contextual information from the MyFitnessPal mobile app and used it to search for an equivalent ingredient. For example, chicken and beef can substitute each other if they are often consumed with rice and salad. \cite{Shirai2020IdentifyingIS} addressed the issue of specific dietary requirements. Similar to our work, they combined the semantic information from a knowledge graph of food and a word embedding model to substitute ingredients to improve a dish's nutritional value. Similar works were done by \cite{Gaillard2015ImprovingIS} using adaptation rules and by \cite{Pellegrini2021ExploitingFE} using the transformer-based model {\sc{BERT}}. However, these works solely explored ingredient substitution, whereas we incorporate this idea for robotic execution of a recipe. Conceptually, our work is similar to \cite{daruna,Mitrevski2021OntologyAssistedGO} that use ontology and domain knowledge to derive missing information. In addition to the knowledge required to manipulate unknown objects, a task tree shows how they can be integrated with other known or unknown items in a complex task like cooking. 

\section{Proposed Method}
\label{sec:task_tree_mod}

Our objective in this work is to generate FOON task plans for never-before-seen recipes through a method that uses FOON as a knowledge base. 
This method requires a recipe's ingredient set $\Ingredients$ and associated dish type $\DishType$ as input and produces a task tree outlining how to prepare the dish.
Due to the limited recipe variety in FOON, the likelihood of encountering recipes with different ingredient combinations is very high.
Specifically, FOON may not contain references to the desired recipe, or, more likely, many ingredients will be absent from FOON.
Despite this shortcoming, our idea is to use the limited knowledge of a FOON to produce task trees for novel recipes, so as to take advantage of existing recipe knowledge bases like Recipe1M+~\cite{marin2019learning}.
The main idea is to retrieve a reference task tree from FOON, which closely resembles the desired recipe, and then modify it to produce a final task tree as shown in Figure~\ref{fig:pipeline3}.
The three main steps of our method are discussed in the following subsections. 



\subsection{Identifying a Reference Goal Node}
\label{sec:task_tree_mod_A}


From a universal FOON, we extract a recipe's task tree by retrieving all functional units related to the final state of a dish, which we refer to as a \textit{goal node}. It is likely that a goal node for a dish does not exist in FOON; in this case, we try to find a suitable node that closely matches the required dish, which we refer to as a \textit{reference goal node} (denoted as $\Goal$). Such a node is used to extract a \textit{reference task tree} $\TaskTree$, which is simply a task tree that creates the reference goal object. 
%
%
%
%
To facilitate the selection of the ideal reference goal node $\Goal$, we created a recipe classification with 30 dish classes (e.g., salad, pizza), with which we can categorize recipes in the form of FOON subgraphs.
The purpose of this classification is to find similar recipes known to us given a dish type. 



%
\begin{algorithm}[ht!]
\caption{Find a reference goal node}
\label{alg:finding_goal}
\begin{algorithmic}[1]
    \Statex \hspace{-2em} \textbf{Input:} given ingredients $\Ingredients$, dish type $\DishType$
    \State $\Candidates$ $\gets$ Retrieve recipes of type $\DishType$ 
    \For{each $\Subgraph$ in $\Candidates$} 
        \State $\SetS \gets$ Ingredients of $\Subgraph$
		\State $\Score \gets$ Find similarity between $\Ingredients$ and $\SetS$
	\EndFor
	\State $\Goal \gets$ End product of subgraph with maximum $\mathsf{score}$
    \Statex \hspace{-2em} \textbf{Output:} $\Goal$
\end{algorithmic}
\end{algorithm}

Algorithm \ref{alg:finding_goal} uses this classification to find candidate goal nodes. For example, if $\DishType = soup$, candidates can be \{\textit{corn soup}, \textit{potato soup}, ...\} depending on the soup recipes that exist in FOON. We store the list of ingredients required to prepare a dish in its goal node so that we can easily identify the required ingredients to make a dish.
Next, the algorithm checks each candidate goal node to select the one that has maximum Word2Vec similarity with the ingredient set $\Ingredients$.
%
%

%
We use a word embedding model provided by spaCy~\cite{spacy}, an open-source library for natural language processing, to get a vectorized representation of each word.
Similar words are grouped together in the embedding space.
For instance, \textit{cucumber} and \textit{zucchini} will be close to each other but far away from \textit{salt}. Similarity is given as a score ranging from 0 to 1, where 1 indicates that two items are semantically similar. In this work, we deem two items similar if the score exceeds a predefined threshold value of 0.90.



    


\subsection{Extracting a Reference Task Tree}

A dish can be prepared in many ways depending on its ingredients and recipe instructions. 
Similarly, in FOON, there are different paths to reach a goal node. We aim to find the path that closely resembles the desired recipe to use it as a base of reference for our final task tree. 
We assume that all utensils and ingredients needed for a recipe are available in the kitchen,  
although it is possible that the ingredients are not available in the desired state. 
For example, a recipe may require \textit{sliced carrot}, but the kitchen contains \textit{whole carrot}. 

\begin{figure}[t]
	\centering
	\includegraphics[width=7cm]{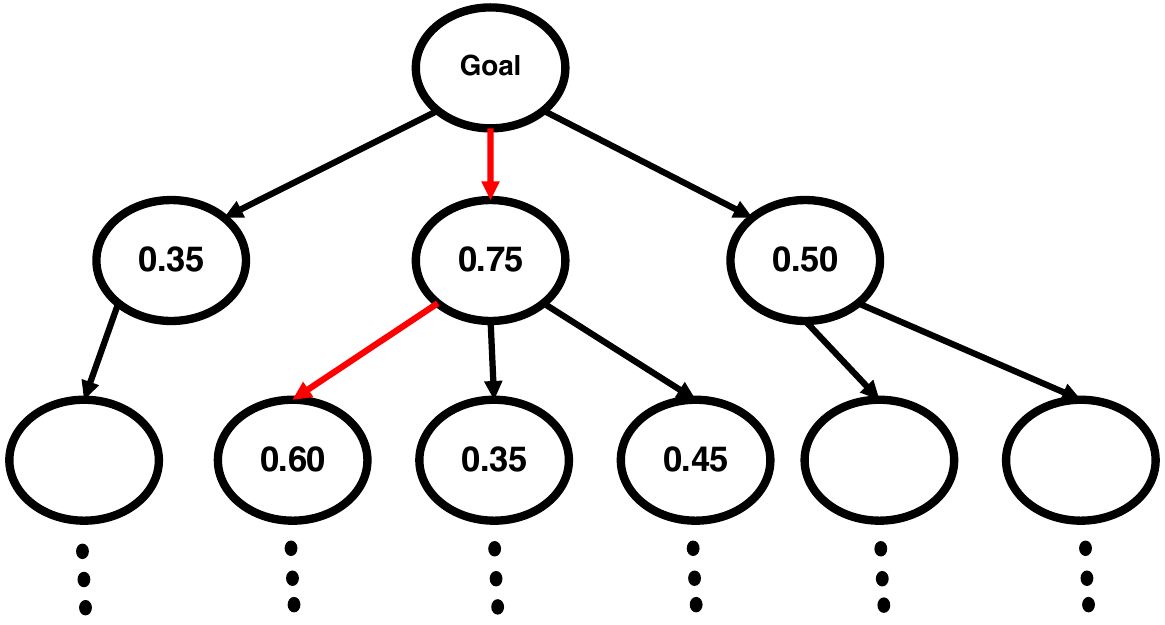}
	\caption{Overview of our reference task tree retrieval method. Each node's value indicate a score returned by the heuristic function; empty nodes are not reached. Red arrows denote the chosen path based on maximum score.}
	\label{fig:heuristic}
\end{figure}
%
%
%
In a typically large FOON, exploring all paths can be time-consuming and memory intensive. Hence, unlike the previous search algorithm that aims to find the most optimal task tree for making a certain object~\cite{paulius2021weighted}, we use \textit{Best First Search}.
At each level, this method evaluates all candidate nodes based on the heuristic function  
    $ g(n) = |A_n \cap \Ingredients|$,
where $A_n$ and $\Ingredients$ denote the set of ingredients stored in node $n$ and ingredients to be added to the dish respectively, and $g(n)$ computes a similarity score checking the overlap between the two sets of ingredients.
We use Word2Vec similarity instead of word matching so that equivalent ingredients are considered as an overlap. 
The similarity score indicates how close we are to the end product (i.e., final dish). 
The node with the maximum score is considered as the closest to the goal and thus will be favored over the other candidates as shown in Figure \ref{fig:heuristic}. 


\begin{algorithm}[ht!]
\caption{Retrieve the task tree of the reference goal node}
\label{alg:reference_retrieval}
\begin{algorithmic}[1]
    \Statex \hspace{-2em} \textbf{Input:} reference goal node $\Goal$ and given ingredients $\Ingredients$ 
    \State $\TaskTree \gets $ A list of functional units in the task tree 
    \State $K$ $\gets$ A list of items available in the kitchen
    \State $Q$ $\gets$ A queue of items to search
    \State $Q$.push($\Goal$)
   
    \While{$Q$ is not empty}
        \State $L \gets$ $Q$.dequeue()
        \If{$L$ does not exist in $K$}
        \State $C \gets$ Find all functional units that create $L$ 
        \State $C^\prime \gets$ Select one from $C$ using heuristic function
        
        \State $\TaskTree$.append($\C^\prime$) 
        
        \For{each input $n$ in $C^\prime$}
            \If{$n$ is not visited}
            \State $Q$.enqueue($n$)
            \State Mark $n$ as visited
            \EndIf
        \EndFor
        \EndIf
        
    \EndWhile
    
    \State $\TaskTree$.reverse() 

    \Statex \hspace{-2em} \textbf{Output:} $\TaskTree$
\end{algorithmic}
\end{algorithm}

The retrieval algorithm (presented as Algorithm~\ref{alg:reference_retrieval}) works as follows.
Let the initial task tree $\TaskTree$ be an empty list. $\TaskTree$ will be populated with the functional units required to prepare the dish defined by the goal node $\Goal$. Let $K$ be the list of items in the kitchen and $Q$ be a queue for keeping track of items that need to be explored. 
At each iteration, we remove the first item $\mathcal{L}$ from $Q$ and check if it is found in $K$ in its desired state. 
If it does not contain $\mathcal{L}$, we search for the functional units $\mathcal{C}$ that create it (i.e., functional units that contain $\mathcal{L}$ as an output node). For instance, while searching for \textit{cake batter}, we may find that it is not available in the kitchen, but there are a few functional units in FOON that show how to make different types of \textit{cake batter} made up of varying sets of ingredients. These units are added to $\mathcal{C}$ as potential candidates. We choose the best candidate $\mathcal{C}^\prime$ from $\mathcal{C}$ based on our heuristic function and add it to the task tree. To execute $\mathcal{C}^\prime$, we have to make sure that we know how to prepare all of its input ingredients. This is why all of its input ingredients are added to $Q$ to be searched next. 
The search continues until $Q$ is empty, meaning that all of the required items are discovered. At the end, $\TaskTree$ is reversed to provide the functional units in the correct sequence that a robot can follow to prepare the dish at $\Goal$.

\subsection{Task Tree Modification}
\label{sec:task-tree-mod}

The reference tree $\TaskTree$ gives a blueprint about how the tasks should be executed, but it may not contain all the ingredients required for a new recipe. Therefore, we need to modify $\TaskTree$ to include missing ingredients. 
The modification procedure first integrates all required ingredients and then removes those that were not given as input. If an ingredient does not exist in FOON, or if it exists in a different state, we find a semantically similar ingredient in the knowledge base using a multi-level search. 
First, we try to find a suitable match for similar ingredients in $\TaskTree$ because we already know how the ingredients in $\TaskTree$ are integrated together to produce the final dish. This gives us high confidence to incorporate missing ingredients. Otherwise, we search in the universal FOON for an equivalent ingredient.
As in Section~\ref{sec:task_tree_mod_A}, we use spaCy's word embedding model to find the closest object in the embedding space. 
Each ingredient contains an object name and its state. 
When checking if an ingredient exists in $\TaskTree$, there are four possible cases depending on the match with the object and state (also presented in Table~\ref{table:substitution_rules}): 
            
%

\textbf{Case 1:}
If the given ingredient exactly matches a leaf node (i.e., a starting node in the state required by the recipe) in $\TaskTree$ in terms of object and state name, no adjustment is required. 

\textbf{Case 2:}
The exact object is not found in $\TaskTree$, but there is an equivalent ingredient with the exact same state. In that case, we can simply substitute the equivalent object name with that of the required object. For instance, if \textit{chopped chili pepper} is required in the recipe but its equivalent in $\TaskTree$ is \textit{chopped jalape\~{n}o}, we can replace \textit{jalape\~{n}o} with \textit{chili pepper}. If \textit{jalape\~{n}o} is also required, a copy of the functional units required for processing \textit{jalape\~{n}o} is created, after which substitution is performed to preserve \textit{jalape\~{n}o} in $\TaskTree$.

\textbf{Case 3:}
If the object is found in $\TaskTree$ but not in the desired state, we search FOON for functional units needed to get the object in the required state. We do this with another retrieval query using Algorithm \ref{alg:reference_retrieval}, where the goal node will be an object node in the leaf of $\TaskTree$. We then add the retrieved functional units to that leaf. For example, if \textit{whole tomato} is given as the starting state but there is \textit{diced tomato} in the leaf of the tree, we will search using the goal node \textit{diced tomato} as the criteria and the added functional units will reflect how \textit{diced tomato} can be made from its \textit{whole} state. 

\textbf{Case 4:}
The object and state of the equivalent ingredient do not match with the given ingredient. We can handle this case by combining the solution from the previous two cases. First, we  substitute the object name and then add the necessary functional units to achieve its desired state. 


\begin{table*}[ht]
\caption{Rules to modify the reference task tree based on the difference between a given ingredient and its equivalent in the tree as discussed in Section~\ref{sec:task-tree-mod}. The term \textit{FU} denotes functional unit.}
\centering
\begin{tabular}{|p{0.03\linewidth}|p{0.12\linewidth} | p{0.14\linewidth} | >{\centering\arraybackslash}p{0.06\linewidth}| >{\centering\arraybackslash}p{0.06\linewidth}|p{0.16\linewidth}|p{0.18\linewidth}|}
\hline
\textbf{Case} & \textbf{End leaf in reference task tree} & \textbf{Given ingredient} & \textbf{Object matched} & \textbf{State matched} & \textbf{Substitution} &\textbf{Add branch} \\ \hline
1 & sliced carrot & sliced carrot & \greencheck & \greencheck & --- & --- \\ \hline
2 & chopped jalape\~{n}o & chopped chili pepper & \redcross &\greencheck & substitute jalape\~{n}o with chili pepper & --- \\ \hline
3 & diced tomato & whole tomato & \greencheck & \redcross & --- & add FUs to make diced tomato from whole tomato\\ \hline
4 & minced shallot & peeled onion & \redcross & \redcross & substitute shallot with onion & add FUs to make minced onion from peeled onion\\ \hline

\end{tabular}
\label{table:substitution_rules}
\vspace{-0.15in}
\end{table*}




\begin{figure}[ht]
    \centering
    \subfloat[FOON-generated task tree for greek salad]{
    {\includegraphics[width=0.95\columnwidth]{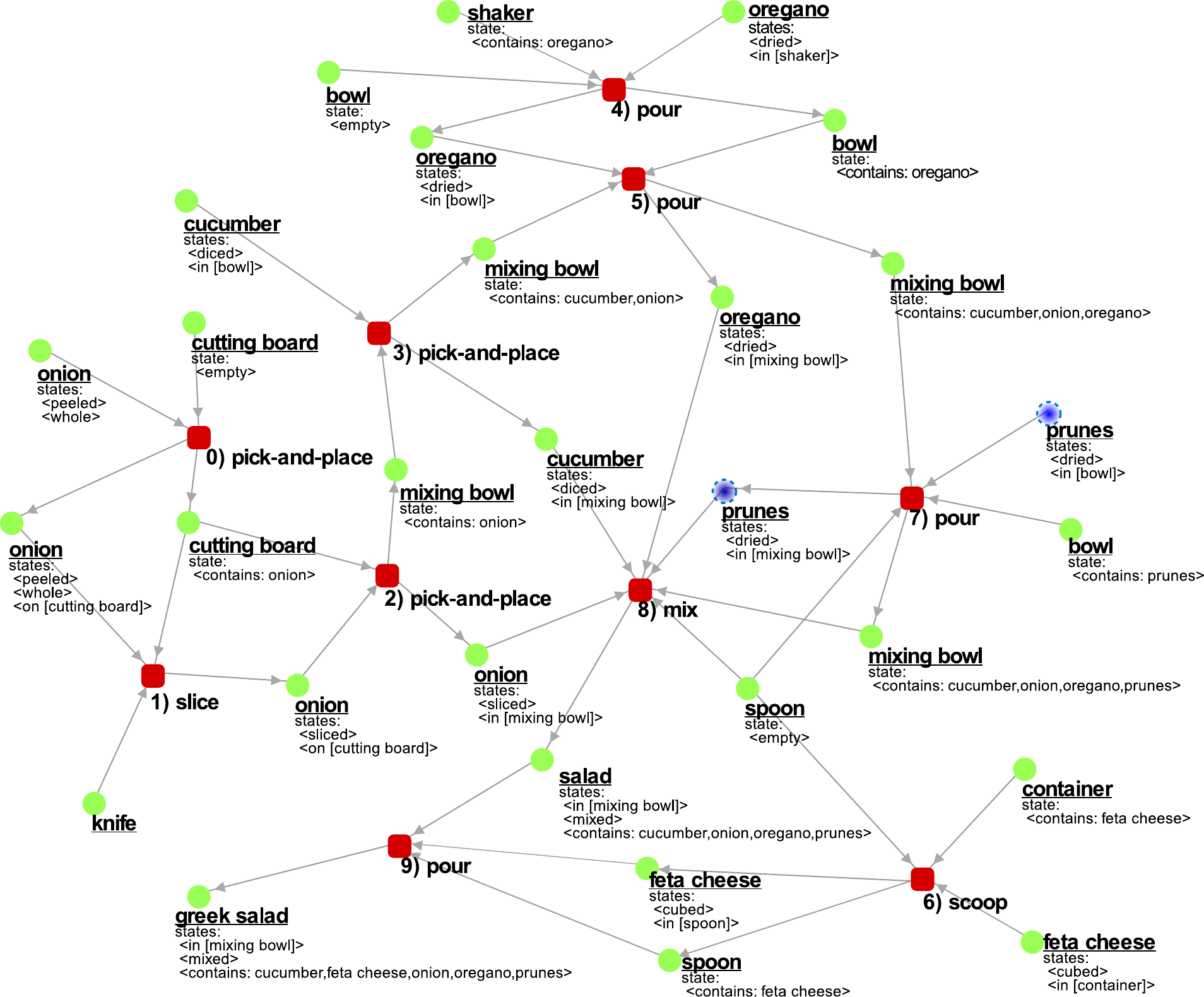}}
    \label{fig:generated-a}}
    
    \subfloat[Progress line for greek salad]{
     {\includegraphics[width=0.95\columnwidth]{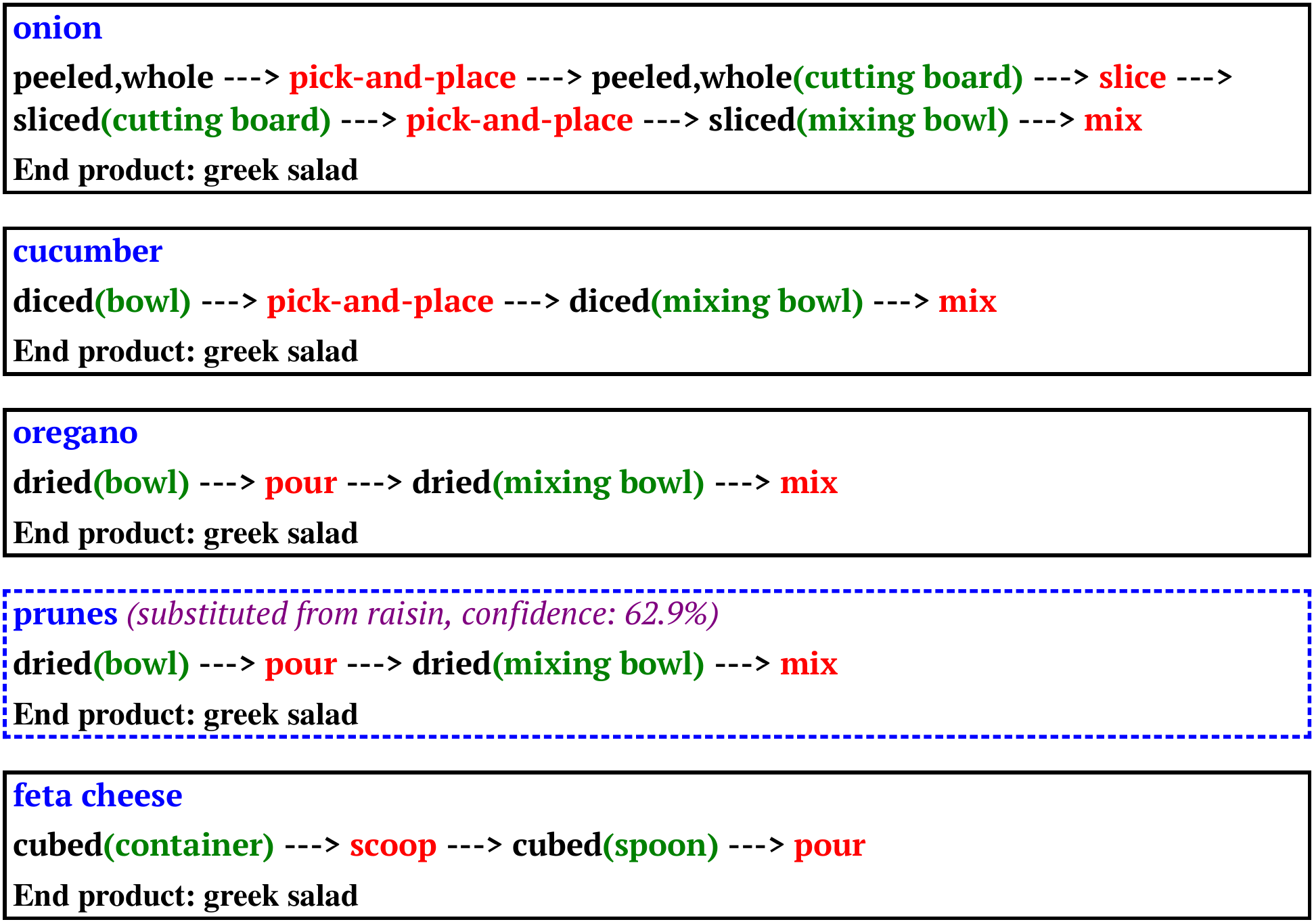}}
     \label{fig:generated-b}}

  \caption{Illustration of: a) a task tree generated with our approach, and b) its corresponding progress line visualization.}
  \label{fig:confusion_matrix}
\end{figure}

If the equivalent ingredient does not exist in the retrieved task tree but is featured in other recipes in FOON, it poses some extra challenges. Let us assume that the recipe requires \textit{roasted turkey} but the closest object to \textit{turkey} available in FOON is \textit{chicken}. Normally, \textit{chicken} can have states such as \textit{raw}, \textit{fried}, \textit{baked}, and \textit{marinated} across recipes. This makes it difficult to choose the state best suited for the task tree. To aid in selecting an appropriate state, we used a state taxonomy~\cite{jelodar2019joint} with 12 different categories, such as \textit{raw}, \textit{finely separated}, and \textit{liquid}.
Similar states, such as \textit{sliced} and \textit{diced}, are grouped together such that existing states can be used as a blueprint for other states of the same category.
If there are multiple candidates to choose from, we break the tie by favoring the most frequently used state in the recipes of the given dish type. 

When substituting a state, we also have to select an appropriate manipulation for which the state is valid. For instance, with the states \textit{diced} and \textit{sliced}, we treat the appropriate action with different motion verbs \textit{dice} and \textit{slice}. For this, we find the most frequently used verb associated with the required state and use it as the motion node label. Once we retrieve the functional units to prepare an ingredient, the challenge is to connect them to the reference tree. To do this, we find functional units in $\TaskTree$ that can easily be modified, which are those that new ingredients can be added as input nodes. For example, any unit that involve mixing or adding ingredients to a container can accept any number of inputs, whereas a robot cannot cut or chop multiple items at the same time. Therefore, for each missing ingredient, we connect the subtrees to these types of functional units, resulting in a connected acyclic graph. We do not need to create this connection if the equivalent ingredient already exists in $\TaskTree$. Finally, the modified task tree will have a list of functional units containing the input ingredients that, if executed sequentially, will produce the desired dish.

\section{Experiments}
\label{sec:experiments}

It is difficult to quantitatively compare a knowledge graph-based method with others, 
as the benefit we get from it lies in its flexibility and adaptability to answer a query. 
Therefore, in our experiments, we used Recipe1M+ to create a diverse set of queries for ingredient combinations from several dish classes as shown in Table~\ref{table:statistics}.
Our focus is on generating a correct task tree for recipes in Recipe1M+.
To make our evaluation more robust, we excluded dish classes for which we had less than 4 recipes in FOON, as it is difficult to gain a general picture of a dish class with too few recipes. Among the selected dish types, we randomly selected 5 classes for our evaluation: \textit{salad, cake, soup, omelette} and \textit{drinks}. Next, 100 recipes of each category were randomly selected from Recipe1M+ for a total of 500 recipes.
We ran these experiments on a MacBook Pro equipped with 2.6 GHz 6-Core Intel Core i7 processor 
and 16GB RAM.



\subsection{Evaluation Strategy}
Our objective is to evaluate the quality of a task tree for unseen ingredient combinations from the Recipe1M+ dataset. 
It is impossible to judge a recipe by solely relying on automated testing such as Intersection over Union (IoU) due to there being many ways to prepare a meal. Furthermore, despite having the same ingredients, one recipe can differ from another recipe in terms of cooking steps while both are valid or correct. 
Hence, we chose to evaluate the FOON-generated task trees by manually checking each task plan. Although it requires a lot of manual effort, we believe that human judgement is necessary to evaluate recipes. To make the process easier, instead of checking the functional units in the task tree, we check each ingredient by reviewing its \textit{progress line}.

\textbf{Progress Line:} In cooking, ingredients undergo several changes in their physical state (e.g., from \textit{whole} to \textit{sliced}) and/or location. We refer to this sequence of changes as an ingredient's \textit{progress line}, which is another visualization of a task tree. We developed a tool to view the progress line of each ingredient to manually evaluate generated task trees. Figures~\ref{fig:generated-a} and~\ref{fig:generated-b} shows an example of a generated task tree and its progress line respectively, where objects, states and motions are colored in black, green and red respectively. 




    

\begin{table}[t]
\centering
\caption{Overview of the test dataset created from Recipe1M+}
\begin{tabular}{|p{0.15\linewidth} | p{0.30\linewidth} | p{0.30\linewidth}|}
\hline
\textbf{Dish Type} & \textbf{No. of Recipes in Recipe1M+}  & \textbf{Average No. of Ingredients per Recipe}  \\ \hline
\textit{salad} & 58,869 & 9  \\ \hline
\textit{cake} & 29,887 & 10 \\ \hline
\textit{soup} & 35,357 & 10  \\ \hline
\textit{omelette} & 686 & 7  \\ \hline
\textit{drinks} & 2,732 & 5 \\ \hline
\end{tabular}
\label{table:statistics}
\vspace{-0.2in}
\end{table}

\subsection{Results}
\begin{figure}[!ht]
	\centering
	\includegraphics[width=\columnwidth]{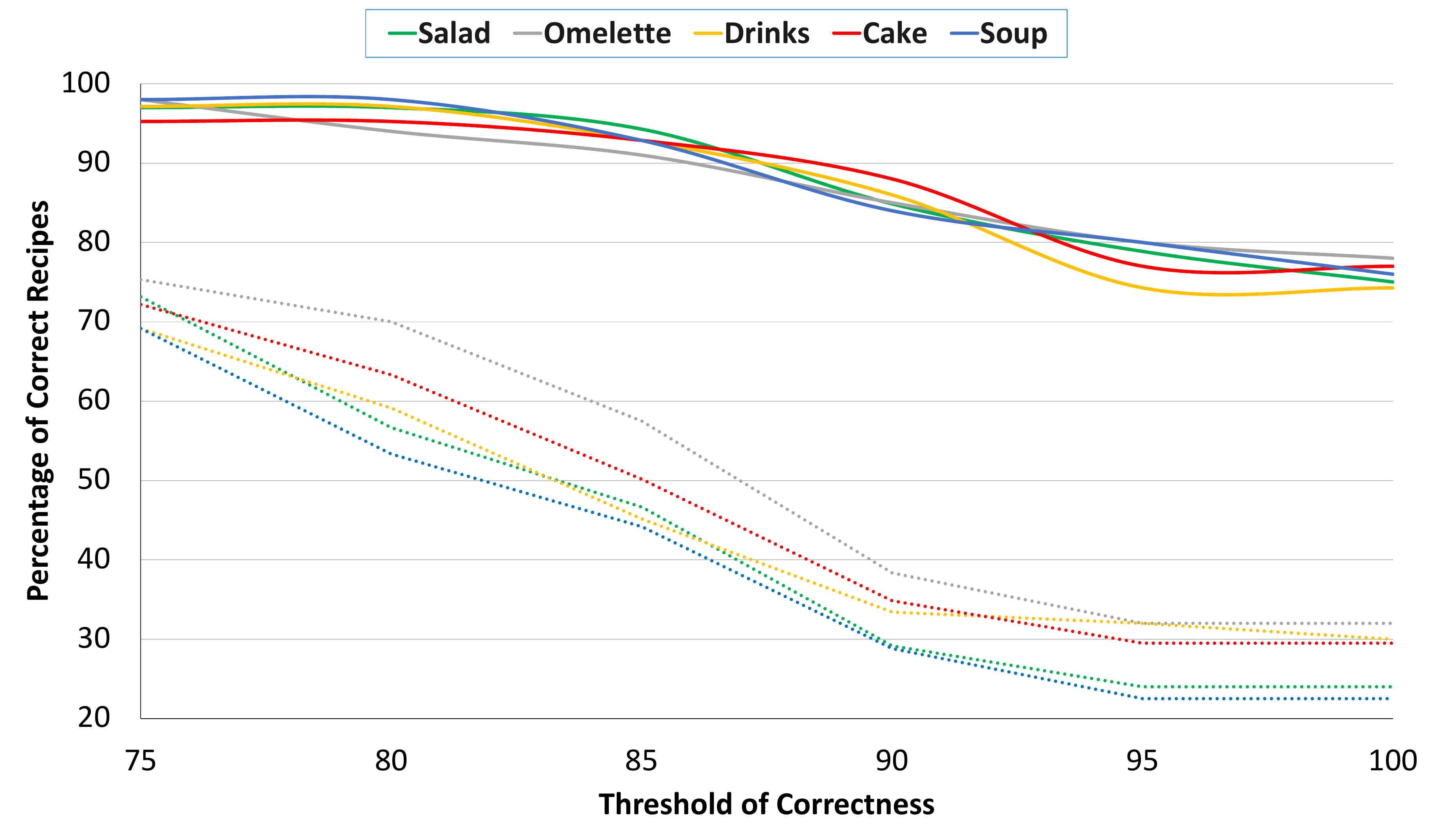}
	\caption{Graph showing the relationship between the threshold of correctness (X-axis) of each recipe and the number of recipes (Y-axis) we can generate with that threshold. The result of recipe generation with and without using task tree modification are presented as solid and dashed lines respectively; the same colored lines indicate the same dish type.} 
	\label{fig:result}
\end{figure}

Task trees generated for the 500 recipes were thoroughly evaluated using the progress line visualizer. When checking a recipe, we review each of its ingredients and indicate whether a given ingredient is \textit{incorrect}, \textit{partially incorrect} or \textit{correct} by assigning it a score from \{0,1,2\}, where `0' means that an ingredient's progress line is incorrect, `1' means partially incorrect, and `2' means correct. A label of {partially incorrect} means that 
a state label assigned to the ingredient is not contextually or physically relevant to it. 
Without modifying the reference task tree, we would only see higher scores for ingredients that already exist within the unmodified tree, while missing ingredients (i.e., items that need to be added to the tree) or extra ingredients (i.e., items that need to be pruned from the tree) will be assigned an incorrect score.
However, our proposed algorithm will be able to generate task plans for unseen ingredients in addition to already existing items while removing extra ingredients. 
%
In Figure~\ref{fig:result}, we present a comparison of these two approaches by reporting the percentage of correctly generated recipes with varying degrees of thresholds. The result shows that the proposed algorithm can generate 76\% of the recipes correctly on average if the threshold of correctness for each recipe is set to 100\%. When the threshold is 100\%, a task tree must have correct progress lines for each ingredient. It is evident from the experiments that using similarity-based ingredient substitution greatly improves the performance and correctness thresholds across all dish types.


\subsection{User Study}
To verify correctness and completeness, we conducted a user study to see whether there is any significant difference between the FOON generated and Recipe1M+ recipes, following the same approach and questionnaires (see Figure \ref{fig:questions}) from a previous study~\cite{sakib2021evaluating}. Since a task tree in its original format is not easy to read and verify for individuals not familiar with FOON, we translated each tree into recipe-like sentences.
In this study, there were 9 participants, where each was given 10 recipes randomly selected from Recipe1M+ and its equivalent generated recipe from our method to assign ratings.
This survey can be accessed by the provided link.\footnote{Link to survey -- \url{https://bit.ly/3wAK909}}
Our hypothesis $H_o$ is that there is no significant difference between FOON-generated and Recipe1M+ recipes.
With the ratings collected from users for questions 4-7,
we performed a null hypothesis test with significance level $\alpha=0.05$ and obtained the $p$-values $0.12$, $0.34$, $0.20$, $0.64$, $0.68$, $0.73$ and $0.35$ respectively.  
Since all the $p$-values are fairly greater than $\alpha$, we cannot reject $H_o$; hence, we can conclude that the generated recipes are equivalent to Recipe1M+ recipes in terms of correctness, completeness and clarity.

\begin{figure}[ht]
    \centering
	\noindent \fbox{\parbox{0.95\columnwidth}
		{\footnotesize
		   \begin{enumerate}
	           \setlength\itemsep{-0.1em}
		       \item Please describe your proficiency level in cooking.
		       \begin{itemize}[label=$\Box$]
		           \setlength\itemsep{-0.15em}
		           \item I have no experience in cooking
		           \item Beginner home cook
		           \item Intermediate home cook
		           \item Advanced home cook
		           \item I have received culinary training
		       \end{itemize}
		       \item If or when you cook, what type of recipes do you use?
		       \begin{itemize}[label=$\Box$]
		           \setlength\itemsep{-0.15em}
		           \item I mostly use recipes that family or friends shared
		           \item I look for recipes online
		           \item I follow recipes from cookbooks
		           \item I only use ingredients I have available
		       \end{itemize}
		       \item Have you made a dish similar to this recipe? 
		       \begin{itemize}[label=$\Box$]
		           \setlength\itemsep{-0.15em}
		           \item I have made this exact dish
		           \item Yes, but I left out some of the ingredients listed
		           \item Yes, but I added some ingredients not listed
		           \item No
		       \end{itemize}
		   \end{enumerate}
		   \begin{enumerate}
		   \setcounter{enumi}{3}
	           \setlength\itemsep{-0.05em}
		       \item Does this recipe seem correct to you?
		       \item Do the steps appear to be in the right order? 
		       \item Are all steps in the recipe correct? 
		   \end{enumerate}
		   \begin{enumerate}
		   \setcounter{enumi}{6}
	           \setlength\itemsep{-0.05em}
		       \item Does each recipe step give enough information in order to complete it?
		       \item Does the recipe skip steps that are obvious to you? 
		   \end{enumerate}
		   \begin{enumerate}
		   \setcounter{enumi}{8}
	           \setlength\itemsep{-0.05em}
		       \item If attempting to follow the given steps, how confident are you that you could make this dish?
		       \item Do the steps in this recipe appear to be clear and easy to follow?
		   \end{enumerate}
		   
	}}
	\caption{Overview of the survey questions used in our user study.}
	\label{fig:questions}
	\vspace{-0.15in}
\end{figure}

%

\section{Discussion}

Recipe1M+ has a significantly higher quantity of recipes and ingredients than FOON, making it very challenging to produce a task plan that perfectly matches the original textual recipe. Nevertheless, even with the limited concepts available in FOON, the flexibility of the task tree generation process allows us to achieve correct results in 76\% of the cases. This mainly depends on the selection of equivalent objects for unseen ingredients, as an incorrect substitution may introduce errors in an ingredient's progress line. For example, the equivalent of \textit{apricot} in FOON is \textit{lemon} according to the word embedding model we used. However, we cannot squeeze an \textit{apricot} to extract its juice as we can with a \textit{lemon}. Therefore, the progress line of \textit{apricot} does not correctly reflect how it should change its state after each manipulation action. 

An ingredient may be used differently across recipes;
for example, \textit{pineapple} is mixed with other fruits while preparing a \textit{fruit salad}, but added as a topping in a \textit{$pizza$}. For this reason, when we add a branch to the task tree, we follow different rules to decide where the missing ingredient should be added. For instance, in \textit{drink} recipes, branches can be added when objects are being poured into a container whereas for \textit{salad}, ingredients are typically mixed in a bowl. 
These concepts are usually not captured in the word embedding model, and rule-based actions are not often sufficient to accurately infer the task plan. Hence, we will focus on improving our methodology through generalization in the future.
We also noted that it is not always possible to include the user-provided state of an ingredient if the kitchen contains it in a non-reversible state, such as \textit{melted cheese} when \textit{diced cheese} is required. We do not mark those as errors since a new batch of ingredients is required to follow user input. Overall, we noted the following errors in our experiments as incorrect substitution: 39\%, incorrect state: 27\%, incorrect motion: 14\% and incorrect integration: 20\%.


\begin{figure}[!t]
	\centering
	\includegraphics[width=\linewidth]{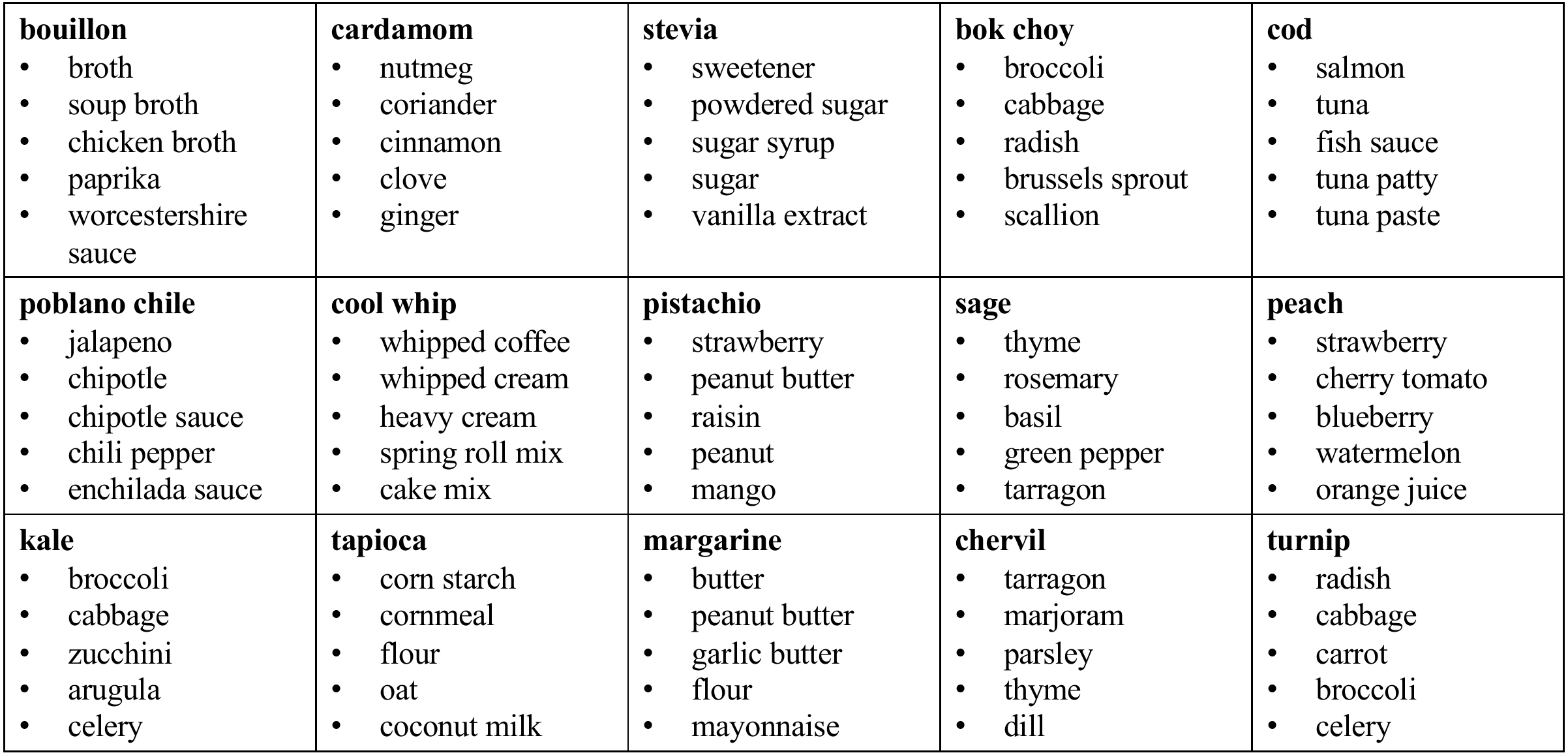}
	\caption{Top 5 equivalent ingredients in FOON for some randomly chosen unseen ingredients from Recipe1M+.}
	\label{fig:substitution}
	\vspace{-0.15in}
\end{figure}

Such errors can be minimized with correction from users, where instead of selecting the top suggestion from the embedding model for each unseen ingredient, users can decide on the ideal substitute among the top 5 candidates.
In the case of Figure~\ref{fig:substitution}, corrections can be made for the ingredients \textit{cool whip} and \textit{pistachio} using the second (\textit{whipped cream}) and fourth (\textit{peanut}) ranked candidates respectively. Furthermore, we can enrich the knowledge base by adding recipes with diverse set of ingredients so that we can make substitutions with higher confidence and with minimal user correction. 

Finally, up to this point, recipes are added to FOON by manually labelling cooking videos, which is a very time-consuming and error-prone process. If we can automate the process of creating subgraphs from  video or text, we could easily expand FOON by adding recipes from various sources.
Using a method such as the task tree generation process can help to reduce the amount of effort required by constructing graphs in a semi-automatic manner, in addition to computer vision approaches for extracting graphs from video~\cite{jelodar2018long}.

A task tree can be used for bootstrapping robot task execution, as it provides a high-level description of a sequence of actions needed to be executed by a robot free from domain-specific robot details or properties (e.g. the number of end-effectors, which end-effector to use for certain actions, motion primitives, etc.). 
For instance, a task tree extracted from a knowledge graph containing instructions of furniture assembly can represent how the raw materials should be connected to build the furnished product.
Although a task tree cannot be executed in its original state, it can however be connected to existing frameworks for manipulation planning, such as PDDL where functional units can be treated as planning operators and can then be used to retrieve plans corresponding to robot low-level motion primitives~\cite{paulius2021roadmap}.

\subsection{Complexity Analysis of the Algorithm}
A FOON is represented as an adjacency list, which allows us to retrieve  functional unit candidates in $\mathcal{O}(1)$ time. 
In the worst case, the number of candidates is equal to the total number of FOON recipes. However, this number is very small on average and thus is negligible. 
Therefore, for $n$ functional units in the task tree, the required time is $\mathcal{O}(n)$.
At the tree modification stage, if an equivalent ingredient exists in the reference tree, substitution can be done in $\mathcal{O}(1)$ time. However, if no suitable equivalent is present, we have to search other recipes for equivalents, which takes $\mathcal{O}(k)$ time, where $k$ is the number of unique ingredients in FOON. We accelerate this process by computing a mapping of equivalent ingredient in the pre-processing phase to find an equivalent ingredient in $\mathcal{O}(1)$ time. Hence, the overall time complexity of the search algorithm is $\mathcal{O}(n)$, where $n$ is the length of the retrieved path. The space complexity is also $\mathcal{O}(n)$ since we store only one path. With the addition of new recipes, FOON increases in size, which in turn can potentially create alternative paths to a goal node. Exploring all paths will incur a larger time and space complexity. 
To avoid that, we chose the heuristic-based search algorithm where the heuristic was designed using the given ingredients. 
Overall, we found the average computation time required to solve a problem instance was 80 milliseconds.


\section{Conclusion and Future Work}
\label{sec:con}


In this work, our goal was to create a system that will enable a robot to derive task plans in novel scenarios not found in its knowledge base. It is almost impossible to equip a robot with knowledge of every possible recipe with countless combinations of ingredients. To address this, we explored recipe generation using FOON as a knowledge base. Similar to how humans can modify a recipe dynamically by using 
their
previous cooking knowledge to determine how items available in the kitchen can be used, we demonstrate how an unseen ingredient can be integrated to an existing recipe by learning from an equivalent ingredient based on semantic similarity metrics. In our experiments, we generated task plans for different ingredient combinations using our approach and showed that plans were accurate in most cases.

The evaluation method used in this work is subjective. In the future, we will explore how generated task trees can be executed with a robot to concretely evaluate our method. Furthermore, we aim to enrich our FOON knowledge base with a more diverse array of recipes by semi-automatically creating subgraphs from instructional videos or text-based sources such as cookbooks.
Furthermore, additional work is required to learn a new embedding model that would not only measure the similarity between two words but also the similarity of two functional units or two task plans that considers the context of the ingredients.

\ifCLASSOPTIONcaptionsoff
  \newpage
\fi



%
\bibliographystyle{IEEEtran}
\bibliography{IEEEabrv,ref}

\begin{thebibliography}{10}
\providecommand{\url}[1]{#1}
\csname url@samestyle\endcsname
\providecommand{\newblock}{\relax}
\providecommand{\bibinfo}[2]{#2}
\providecommand{\BIBentrySTDinterwordspacing}{\spaceskip=0pt\relax}
\providecommand{\BIBentryALTinterwordstretchfactor}{4}
\providecommand{\BIBentryALTinterwordspacing}{\spaceskip=\fontdimen2\font plus
\BIBentryALTinterwordstretchfactor\fontdimen3\font minus
  \fontdimen4\font\relax}
\providecommand{\BIBforeignlanguage}[2]{{%
\expandafter\ifx\csname l@#1\endcsname\relax
\typeout{** WARNING: IEEEtran.bst: No hyphenation pattern has been}%
\typeout{** loaded for the language `#1'. Using the pattern for}%
\typeout{** the default language instead.}%
\else
\language=\csname l@#1\endcsname
\fi
#2}}
\providecommand{\BIBdecl}{\relax}
\BIBdecl

\bibitem{jelodar2019joint}
A.~B. Jelodar and Y.~Sun, ``{Joint Object and State Recognition using Language
  Knowledge},'' in \emph{2019 IEEE International Conference on Image Processing
  (ICIP)}.\hskip 1em plus 0.5em minus 0.4em\relax IEEE, 2019.

\bibitem{paulius2016functional}
D.~Paulius, Y.~Huang, R.~Milton, W.~D. Buchanan, J.~Sam, and Y.~Sun,
  ``{Functional Object-Oriented Network for Manipulation Learning},'' in
  \emph{Intelligent Robots and Systems (IROS), 2016 IEEE/RSJ International
  Conference on}.\hskip 1em plus 0.5em minus 0.4em\relax IEEE, 2016, pp.
  2655--2662.

\bibitem{Ren2013}
S.~Ren and Y.~Sun, ``Human-object-object-interaction affordance,'' in
  \emph{Workshop on Robot Vision}, 2013.

\bibitem{SunRAS2013}
Y.~Sun, S.~Ren, and Y.~Lin, ``Object-object interaction affordance learning,''
  \emph{Robotics and Autonomous Systems}, 2013.

\bibitem{Lin2015a}
Y.~Lin and Y.~Sun, ``Robot grasp planning based on demonstrated grasp
  strategies,,'' \emph{Intl. Journal of Robotics Research}, vol.~34, no.~1, pp.
  26--42, 2015.

\bibitem{paulius2018functional}
D.~Paulius, A.~B. Jelodar, and Y.~Sun, ``{Functional Object-Oriented Network:
  Construction \& Expansion},'' in \emph{2018 IEEE International Conference on
  Robotics and Automation (ICRA)}.\hskip 1em plus 0.5em minus 0.4em\relax IEEE,
  2018, pp. 5935--5941.

\bibitem{marin2019learning}
J.~Marin, A.~Biswas, F.~Ofli, N.~Hynes, A.~Salvador, Y.~Aytar, I.~Weber, and
  A.~Torralba, ``{Recipe1M+: A Dataset for Learning Cross-Modal Embeddings for
  Cooking Recipes and Food Images},'' \emph{{IEEE} Trans. Pattern Anal. Mach.
  Intell.}, 2019.

\bibitem{Gibson_1977}
J.~Gibson, ``The theory of affordances,'' in \emph{Perceiving, Acting and
  Knowing}, R.~Shaw and J.~Bransford, Eds.\hskip 1em plus 0.5em minus
  0.4em\relax Hillsdale, NJ: Erlbaum, 1977.

\bibitem{mcdermott1998pddl}
D.~McDermott, M.~Ghallab, A.~Howe, C.~Knoblock, A.~Ram, M.~Veloso, D.~Weld, and
  D.~Wilkins, ``{PDDL -- The Planning Domain Definition Language},'' CVC
  TR-98-003/DCS TR-1165, Yale Center for Computational Vision and Control,
  Tech. Rep., 1998.

\bibitem{jelodar2018long}
A.~B. {Jelodar}, D.~{Paulius}, and Y.~{Sun}, ``{Long Activity Video
  Understanding Using Functional Object-Oriented Network},'' \emph{IEEE
  Transactions on Multimedia}, vol.~21, no.~7, pp. 1813--1824, July 2019.

\bibitem{caba2015activitynet}
B.~G. Fabian Caba~Heilbron, Victor~Escorcia and J.~C. Niebles, ``Activitynet: A
  large-scale video benchmark for human activity understanding,'' in
  \emph{Proceedings of the IEEE Conference on Computer Vision and Pattern
  Recognition}, 2015, pp. 961--970.

\bibitem{Damen2018EPICKITCHENS}
D.~Damen, H.~Doughty, G.~M. Farinella, S.~Fidler, A.~Furnari, E.~Kazakos,
  D.~Moltisanti, J.~Munro, T.~Perrett, W.~Price, and M.~Wray, ``{Scaling
  Egocentric Vision: The EPIC-KITCHENS Dataset},'' in \emph{European Conference
  on Computer Vision (ECCV)}, 2018.

\bibitem{foonet}
``{FOON Website:} {Graph Viewer} and {Videos},'' \url{http://www.foonets.com},
  2021, accessed: 2022-07-10.

\bibitem{paulius2021weighted}
D.~Paulius, K.~S.~P. Dong, and Y.~Sun, ``{Task Planning with a Weighted
  Functional Object-Oriented Network},'' in \emph{2021 IEEE International
  Conference on Robotics and Automation (ICRA)}.\hskip 1em plus 0.5em minus
  0.4em\relax IEEE, 2021.

\bibitem{beetz2009}
M.~Tenorth and M.~Beetz, ``{KNOWROB -- knowledge processing for autonomous
  personal robots},'' in \emph{2009 IEEE/RSJ International Conference on
  Intelligent Robots and Systems}, 2009, pp. 4261--4266.

\bibitem{pancake}
M.~Beetz, U.~Klank, I.~Kresse, A.~Maldonado, L.~Mösenlechner, D.~Pangercic,
  T.~Rühr, and M.~Tenorth, ``{Robotic roommates making pancakes},'' in
  \emph{2011 11th IEEE-RAS International Conference on Humanoid Robots}, 2011,
  pp. 529--536.

\bibitem{ikeabot}
R.~A. Knepper, T.~Layton, J.~Romanishin, and D.~Rus, ``{IkeaBot: An autonomous
  multi-robot coordinated furniture assembly system},'' in \emph{2013 IEEE
  International Conference on Robotics and Automation}, 2013, pp. 855--862.

\bibitem{Lee2021IKEAFA}
Y.~Lee, E.~Hu, Z.~Yang, A.~C. Yin, and J.~J. Lim, ``{IKEA Furniture Assembly
  Environment for Long-Horizon Complex Manipulation Tasks},'' \emph{2021 IEEE
  International Conference on Robotics and Automation (ICRA)}, pp. 6343--6349,
  2021.

\bibitem{onto_ikea}
A.~Vassiliades, N.~Zarkadas, N.~Bassiliades, and T.~Patkos, ``{Onto-IKEA: A
  Knowledge Retrieval Framework Based on IKEA Ontology},'' in \emph{JOWO},
  2021.

\bibitem{yang2014manipulation}
Y.~Yang, A.~Guha, C.~Fermuller, and Y.~Aloimonos, ``Manipulation action tree
  bank: A knowledge resource for humanoids,'' in \emph{Humanoid Robots
  (Humanoids), 2014 14th IEEE-RAS International Conference on}.\hskip 1em plus
  0.5em minus 0.4em\relax IEEE, 2014, pp. 987--992.

\bibitem{Sera2021AssemblyPB}
I.~Sera, N.~Yamanobe, I.~G. Ramirez-Alpizar, Z.~Wang, W.~Wan, and K.~Harada,
  ``Assembly planning by recognizing a graphical instruction manual,''
  \emph{2021 IEEE/RSJ International Conference on Intelligent Robots and
  Systems (IROS)}, pp. 3138--3145, 2021.

\bibitem{Varshney2013ABD}
L.~R. Varshney, F.~Pinel, K.~R. Varshney, D.~Bhattacharjya,
  A.~Sch{\"o}rgendorfer, and Y.-M. Chee, ``A big data approach to computational
  creativity,'' \emph{ArXiv}, vol. abs/1311.1213, 2013.

\bibitem{Mller2017CookingME}
G.~M{\"u}ller and R.~Bergmann, ``{Cooking Made Easy: On a Novel Approach to
  Complexity-Aware Recipe Generation},'' in \emph{ICCBR}, 2017.

\bibitem{Jabeen2019EvoChefSM}
H.~Jabeen, N.~Tahara, and J.~Lehmann, ``{EvoChef: Show Me What to Cook!
  Artificial Evolution of Culinary Arts},'' in \emph{EvoMUSART}, 2019.

\bibitem{Jabeen2020AutoChefAG}
H.~Jabeen, J.~Weinz, and J.~Lehmann, ``{AutoChef: Automated Generation of
  Cooking Recipes},'' \emph{2020 IEEE Congress on Evolutionary Computation
  (CEC)}, pp. 1--7, 2020.

\bibitem{lida1}
K.~Junge, J.~Hughes, T.~George~Thuruthel, and F.~Iida, ``Improving robotic
  cooking using batch bayesian optimization,'' \emph{IEEE Robotics and
  Automation Letters}, vol.~PP, pp. 1--1, 01 2020.

\bibitem{DBLP:journals/corr/AchananuparpW16}
\BIBentryALTinterwordspacing
P.~Achananuparp and I.~Weber, ``{Extracting Food Substitutes From Food Diary
  via Distributional Similarity},'' \emph{CoRR}, vol. abs/1607.08807, 2016.
  [Online]. Available: \url{http://arxiv.org/abs/1607.08807}
\BIBentrySTDinterwordspacing

\bibitem{Shirai2020IdentifyingIS}
S.~Shirai, O.~W. Seneviratne, M.~Gordon, C.-H. Chen, and D.~L. McGuinness,
  ``Identifying ingredient substitutions using a knowledge graph of food,''
  \emph{Frontiers in Artificial Intelligence}, vol.~3, 2020.

\bibitem{Gaillard2015ImprovingIS}
E.~Gaillard, J.~Lieber, and E.~Nauer, ``{Improving Ingredient Substitution
  using Formal Concept Analysis and Adaptation of Ingredient Quantities with
  Mixed Linear Optimization},'' in \emph{ICCBR}, 2015.

\bibitem{Pellegrini2021ExploitingFE}
C.~Pellegrini, E.~{\"O}zsoy, M.~Wintergerst, and G.~Groh, ``Exploiting food
  embeddings for ingredient substitution,'' in \emph{HEALTHINF}, 2021.

\bibitem{daruna}
A.~A. Daruna, L.~Nair, W.~Liu, and S.~Chernova, ``Towards robust one-shot task
  execution using knowledge graph embeddings,'' \emph{2021 IEEE International
  Conference on Robotics and Automation (ICRA)}, pp. 11\,118--11\,124, 2021.

\bibitem{Mitrevski2021OntologyAssistedGO}
A.~Mitrevski, P.-G. Pl{\"o}ger, and G.~Lakemeyer, ``{Ontology-Assisted
  Generalisation of Robot Action Execution Knowledge},'' \emph{2021 IEEE/RSJ
  International Conference on Intelligent Robots and Systems (IROS)}, pp.
  6763--6770, 2021.

\bibitem{spacy}
\BIBentryALTinterwordspacing
M.~Honnibal, I.~Montani, S.~Van~Landeghem, and A.~Boyd, ``{spaCy:
  Industrial-strength Natural Language Processing in Python},'' 2020. [Online].
  Available: \url{https://doi.org/10.5281/zenodo.1212303}
\BIBentrySTDinterwordspacing

\bibitem{sakib2021evaluating}
M.~S. Sakib, H.~Baez, D.~Paulius, and Y.~Sun, ``{Evaluating Recipes Generated
  from Functional Object-Oriented Network},'' \emph{arXiv preprint
  arXiv:2106.00728}, 2021, (Featured in \textit{18th International Conference
  on Ubiquitous Robots (UR 2021))}.

\bibitem{paulius2021roadmap}
D.~Paulius, A.~Agostini, Y.~Sun, and D.~Lee, ``{A Road-map to Robot Task
  Execution with the Functional Object-Oriented Network},'' \emph{arXiv
  preprint arXiv:2106.00158}, 2021, (Featured in \textit{18th International
  Conference on Ubiquitous Robots (UR 2021))}.

\end{thebibliography}

%








\end{document}